\documentclass[journal,twoside]{IEEEtran}

\usepackage[noadjust]{cite}
\usepackage{color,soul}
\soulregister\cite7 % 针对\cite命令
\soulregister\citep7 % 针对\citep命令
\soulregister\citet7 % 针对\citet命令
\soulregister\ref7 % 针对\ref命令
\soulregister\pageref7 % 针对\pageref命令

\ifCLASSINFOpdf
  \usepackage[pdftex]{graphicx}
  \graphicspath{{./pdf/}{./jpeg/}}
  \DeclareGraphicsExtensions{.pdf,.jpeg,.PNG}
\else
  \usepackage[dvips]{graphicx}
  \graphicspath{{./eps/}}
  \DeclareGraphicsExtensions{.eps}
\fi
\usepackage{xcolor}
\usepackage[switch]{lineno}
\usepackage{amsmath,amssymb,amsthm,mathtools}
\usepackage{algorithm}
\usepackage{algorithmic}
\usepackage{bm}
\usepackage{multirow}

\usepackage{makecell}

\begin{document}
\title{
Prismatic-Bending Transformable (PBT) Joint for a Modular, Foldable Manipulator with Enhanced Reachability and Dexterity} 
%


\author{Jianshu Zhou$^{1}$, \IEEEmembership{Member,~IEEE,~ASME},
        Junda Huang$^{2}$, 
        Boyuan Liang$^{1}$,\IEEEmembership{ Student Member,~IEEE}, \\
        Xiang Zhang$^{3}$,
        Xin Ma$^{2}$,
        Masayoshi Tomizuka$^{1}$, \IEEEmembership{Life Fellow,~IEEE,~ASME} 
        
\thanks{} 
\thanks{Corresponding to Masayoshi Tomizuka: tomizuka@berkeley.edu}
\thanks{$^{1}$Jianshu Zhou, Boyuan Liang, and Masayoshi Tomizuka are with the Department of Mechanical Engineering, University of California, Berkeley.}
\thanks{$^{2}$Junda Huang and Xin Ma are with The Department of Mechanical and Automation Engineering, The Chinese University of Hong Kong.}
\thanks{$^{3}$Xiang Zhang is with FANUC Advanced Research Laboratory, FANUC America Corporation,
USA.}
}


% \markboth{IEEE TRANSACTIONS ON ...}%
% {\MakeLowercase{\textit{et al.}}: ...}

\maketitle

%\pagestyle{empty}  % no page number for the second and the later pages
%\thispagestyle{empty} % no page number for the first page

\begin{abstract}

Robotic manipulators, traditionally designed with classical joint-link articulated structures, excel in industrial applications but face challenges in human-centered and general-purpose tasks requiring greater dexterity and adaptability. To address these challenges, we propose the Prismatic-Bending Transformable (PBT) Joint—a novel, scissors-inspired mechanism with directional maintenance capability that provides bending, rotation, and elongation/contraction within a single module. This design enables transformable kinematic chains that are modular, reconfigurable, and scalable for diverse tasks. We detail the mechanical design, optimization, kinematic and dynamic modeling, and experimental validation of the PBT joint, demonstrating its integration into foldable, modular robotic manipulators. The PBT joint functions as a single stock keeping unit (SKU), enabling manipulators to be constructed entirely from standardized PBT joints. It also serves as a modular extension for existing systems, such as wrist modules, streamlining design, deployment, transportation, and maintenance. Three joint sizes have been developed and tested, showcasing enhanced dexterity, reachability, and adaptability, particularly in confined and cluttered spaces. This work presents a promising approach to robotic manipulator development, providing a compact and versatile solution for operation in dynamic and constrained environments.

\end{abstract}


% \fi
%\IEEEpeerreviewmaketitle

\section{Introduction}

\begin{figure}[!t] % fig_1
    \centering
    \includegraphics[width=\linewidth]{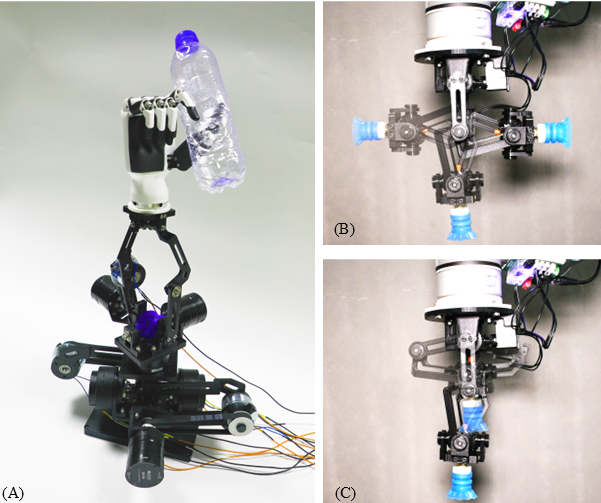}
    \caption{(A) The modular manipulator composed of two PBT joints and a dexterous end-effector. (B) and (C) show the PBT wrist. The PBT wrist is a small-sized PBT joint with a suction cup at its tip. (B) is performing the revolute motion. (C) demonstrates the prismatic motion.}
    \label{fig:intro}
\end{figure}

\IEEEPARstart{R}{}obotic manipulators are essential tools that enable robots to interact with the physical world, much like human arms facilitate our own interactions. Traditional manipulators are typically constructed using two fundamental joint types—prismatic and revolute—connected by rigid links to form various kinematic chains \cite{siciliano2008springer}. These have led to the development of six well-known geometric configurations: articulated, spherical, SCARA, cylindrical, Cartesian, and parallel manipulators \cite{lynch2017modern}. Among these, articulated manipulators are the most widely adopted due to their intuitive kinematics and suitability for high-accuracy, high-payload tasks \cite{horowitz1986adaptive, yao1996smooth}.

However, the increasing demand for robots in human-centered environments, such as domestic, assistive, or service applications, introduces new challenges that emphasize adaptability, dexterity, obstacle avoidance, and reachability over traditional metrics like precision and payload capacity \cite{billard2019trends, okamura2000dexterous, bicchi2000hands, zhou2024dexterous}. Moreover, while robots operating in structured settings can often focus on avoiding obstacles only at the end-effector, those deployed in unstructured environments must consider the full-body configuration of the manipulator in response to dynamic surroundings. Achieving greater flexibility in body posture while maintaining end-effector manipulability is therefore essential for effective manipulation \cite{lynch2017modern, billard2019trends, tomizuka2002mechatronics}.

To address the challenges of dexterity and adaptability in unstructured and human-centered environments, researchers have primarily explored two innovative design paradigms: continuum manipulators and foldable robotic mechanisms. Continuum manipulators, inspired by biological systems such as elephant trunks and octopus arms, enable smooth and adaptive motion through continuously deformable bodies or tendon-driven structures \cite{rus2015soft, russo2023continuum}. These systems provide inherent compliance and multi-directional bending, making them especially well-suited for navigating complex, cluttered environments \cite{gong2021soft, laschi2012soft}. Recent developments in this field have focused on enhancing modeling, control strategies, and task-specific designs for manipulation in constrained and irregular spaces \cite{gong2021soft}. Our previous work has contributed to this area through the development of a tendon-driven soft wrist with human-like articulation for teleoperated nasopharyngeal swab sampling \cite{zhou2021bis, chen2021toos}. We also introduced a tendon-jamming mechanism to achieve tunable stiffness, allowing the structure to adapt to varying load conditions while maintaining dexterous motion \cite{zhou2022bioinspired}.

In parallel, foldable manipulators have emerged as a compelling solution for achieving reconfigurability, compact deployment, and structural adaptability \cite{vincent2000deployable}. These systems utilize mechanisms capable of extending, folding, or reshaping to optimize workspace usage and enhance maneuverability in constrained environments. They are particularly valuable in domains such as disaster response, mobile robotics, and space exploration \cite{kresling2000coupled}. Recent developments have highlighted innovative applications of origami-inspired designs—for example, Kim et al. proposed a self-locking robotic arm that folds flat and deploys into a functional manipulator, achieving both portability and structural rigidity \cite{kim2018origami}. Similarly, Chen et al. introduced ReachBot, a robot designed for extraterrestrial missions that integrates locomotion and manipulation through deployable booms, enabling interaction with distant or vertical surfaces \cite{chen2024locomotion}. Suthar and Jung developed a foldable robot arm for drones using a twisted string actuator, demonstrating a lightweight and compact solution well-suited for aerial manipulation tasks \cite{suthar2021frad}. Complementary to these approaches, Dai et al. have advanced the field of metamorphic and reconfigurable mechanisms, exploring SLPM-based deployable robots and modular systems capable of changing their kinematic structure to meet varying task demands \cite{dai2006metamorphic, gao2019design}.

Despite these advancements, most existing designs fall short in simultaneously providing post-deployment dexterity and scalable modularity. Continuum systems often face challenges in precise control and complex actuation schemes, while foldable systems typically lack adaptability once deployed. Moreover, few platforms successfully integrate multi-directional bending with structural transformability in a single unified architecture—an integration that could significantly enhance the versatility and functional performance of robotic manipulators operating in dynamic and unstructured environments.

To bridge this gap, we propose the Prismatic-Bending Transformable (PBT) Joint—a novel, scissor-inspired mechanism that combines the multi-directional flexibility and foldability. The joint supports bending, rotational, and prismatic motions, enabling smooth transitions between prismatic and bending configurations. This architecture facilitates compact, reconfigurable manipulators that expand into highly dexterous structures when deployed. We present the joint’s design, along with its kinematic and dynamic modeling, and validate its performance through experimental evaluation. Quantitative analysis and comparisons demonstrates improvements in dexterity, manipulability, and workspace reach. Designed as a standardized SKU, the PBT Joint enables scalable manipulator construction without the need for custom components, and it can also function as a modular extension, such as a reconfigurable wrist. We developed and tested three joint sizes—large, medium, and small—across multiple configurations, confirming the joint’s scalability, versatility, and effectiveness for general-purpose manipulation in dynamic, human-centered environments.

The contributions of this work are summarized in three parts:
\begin{itemize}
    \item Development of the Prismatic-Bending Transformable (PBT) Joint, a modular unit integrating multi-directional bending, rotation, and elongation/contraction with a 3D direction maintenance mechanism, offering adaptable dexterity and task-specific customization.

    \item Development of a modular and foldable robotic manipulator architecture using PBT joints, enabling enhanced dexterity, manipulability, and obstacle avoidance, with applications as an extension wrist or a foldable two-joint manipulator for unstructured environments.

     \item Comprehensive analysis, including kinematic, manipulability, workspace, singularity avoidance and dynamic modeling, guides PBT joint and its enabled manipulator design.
\end{itemize}

The paper is organized as follows: Section~\ref{sec:concept} introduces the concept of the PBT Joint. Section~\ref{sec:design} describes the joint’s design and its integration into modular manipulators. Section~\ref{sec:kinematics} analyzes the kinematics, dynamics, and manipulability. Section~\ref{sec:experiments} presents experimental validations for both single joints and modular systems. The paper concludes with a summary and discussion of future work.

\begin{figure}[!t] % fig_1
    \centering
    \includegraphics[width=\linewidth]{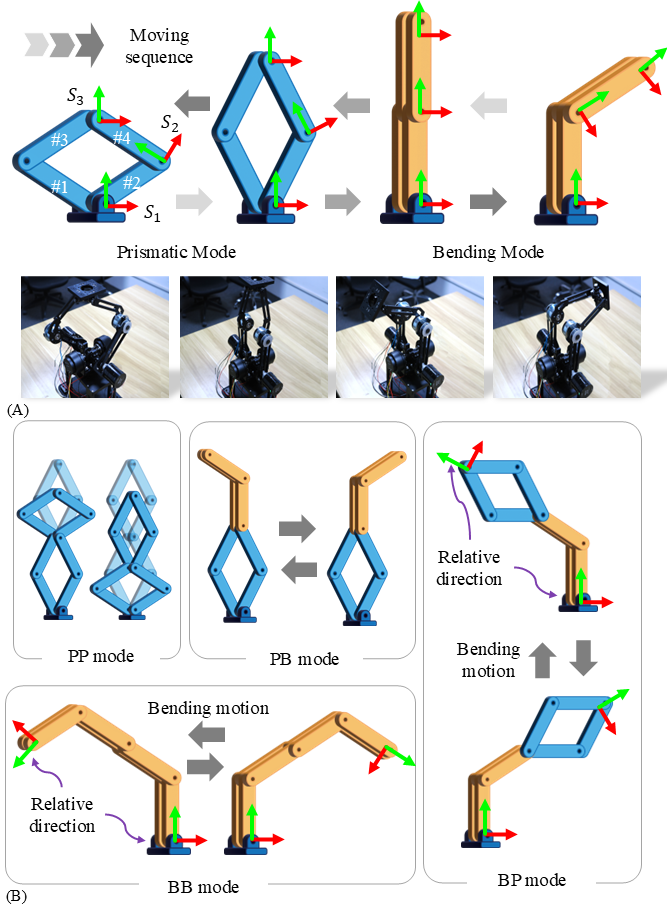}
    \caption{Concept of the Prismatic-Bending transformable (PBT) joint. (A) The motion illustration of the PBT joint. The first three stages are linear motion, while the third and fourth stages are revolute motion. The desired joint angles in the moving process are shown by the coordinates $S_1, S_2, and S_3$. (B) Serial combination of two PBT joints. The 'P' represents prismatic, while 'B' represents bending.}
    \label{fig:Concept}
\end{figure}

\section{Concept of PBT Joint}
\label{sec:concept}

This section introduces the motion and degrees of freedom (DOFs) of a single PBT joint and its modular extension. As illustrated in Figure \ref{fig:Concept}A, the proposed PBT joint employs a scissors-inspired mechanism to achieve reliable prismatic and bending motion, providing two correlated DOFs \cite{luo2017design, yu2021versatile, he2019mechanical}. This mechanism forms the basic PBT joint, enabling prismatic and bending transformations. Self-rotation is achieved by integrating a perpendicular rotation actuator with the core scissors mechanism \cite{maden2011review}.

Specifically, assuming the base frame (\(\text{S}_1\)) is fixed, driving Links \#1 and \#2 near the base causes the distal endpoint (\(\text{S}_3\)) to perform linear motion within the range \((0, 2L]\). When \(\text{S}_3\) reaches the farthest point, the PBT structure transitions to the serial chain, which is similar to a traditional revolute joint. At this stage, applying torque to the $S_2$ enables rotational motion between the coincident Links \#3, \#4 and Links \#1, \#2.

For modular extension, Figure \ref{fig:Concept}B illustrates two serially connected PBT joints, enabling four transformable kinematic modes: (i) prismatic-prismatic (PP) motion, (ii) prismatic-bending (PB) motion, (iii) bending-prismatic (BP) motion, and (iv) bending-bending (BB) motion. Each stage operates independently, allowing versatile motion configurations.

\begin{figure}[!t] % fig_1
    \centering
    \includegraphics[width=\linewidth]{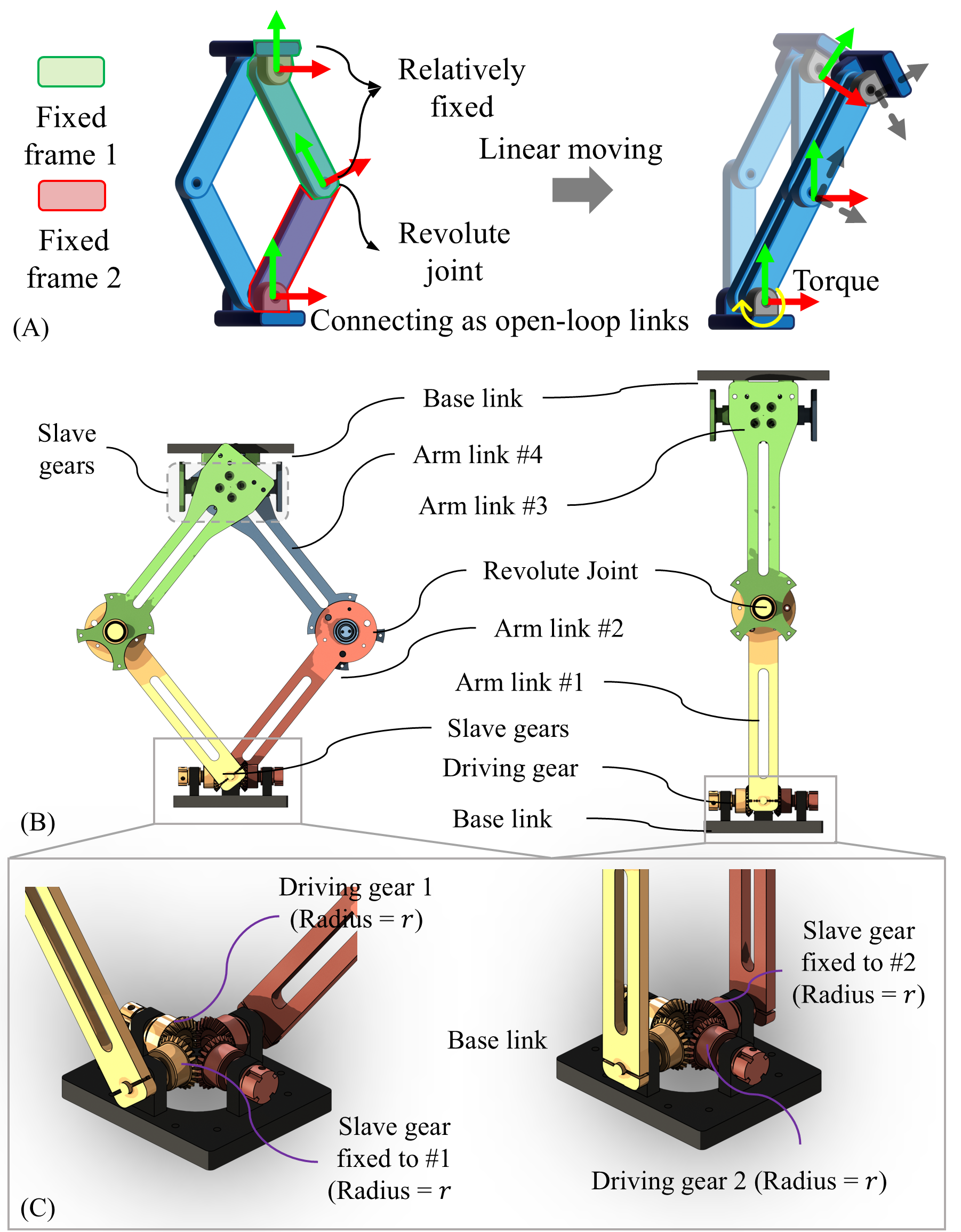}
    \caption{3D Direction Maintenance Mechanism. (A) Illustrates the Direction Maintenance Challenge in the PBT Joint. Using an open-loop link connection to implement the PBT joint causes the direction change at the PBT joint's end. (B) PBT Joint with 3D Direction Maintenance Mechanism: Each end of the base links is equipped with a direction maintenance structure. (C) 3D Direction Maintenance Structure: The mechanism effectively withstands overturning torque across all three axes.}
    \label{fig:direction_maintain}
\end{figure}

\section{Design of PBT Joint and Modular Manipulator Architecture}
\label{sec:design}

In this section, we present the detailed design of the PBT joint and its resulting modular manipulator. This includes the direction-maintenance mechanism of the PBT joint, the design and actuation principles, and the assembly process of the modular manipulator.

\begin{figure}[!t] % fig_1
    \centering
    \includegraphics[width=\linewidth]{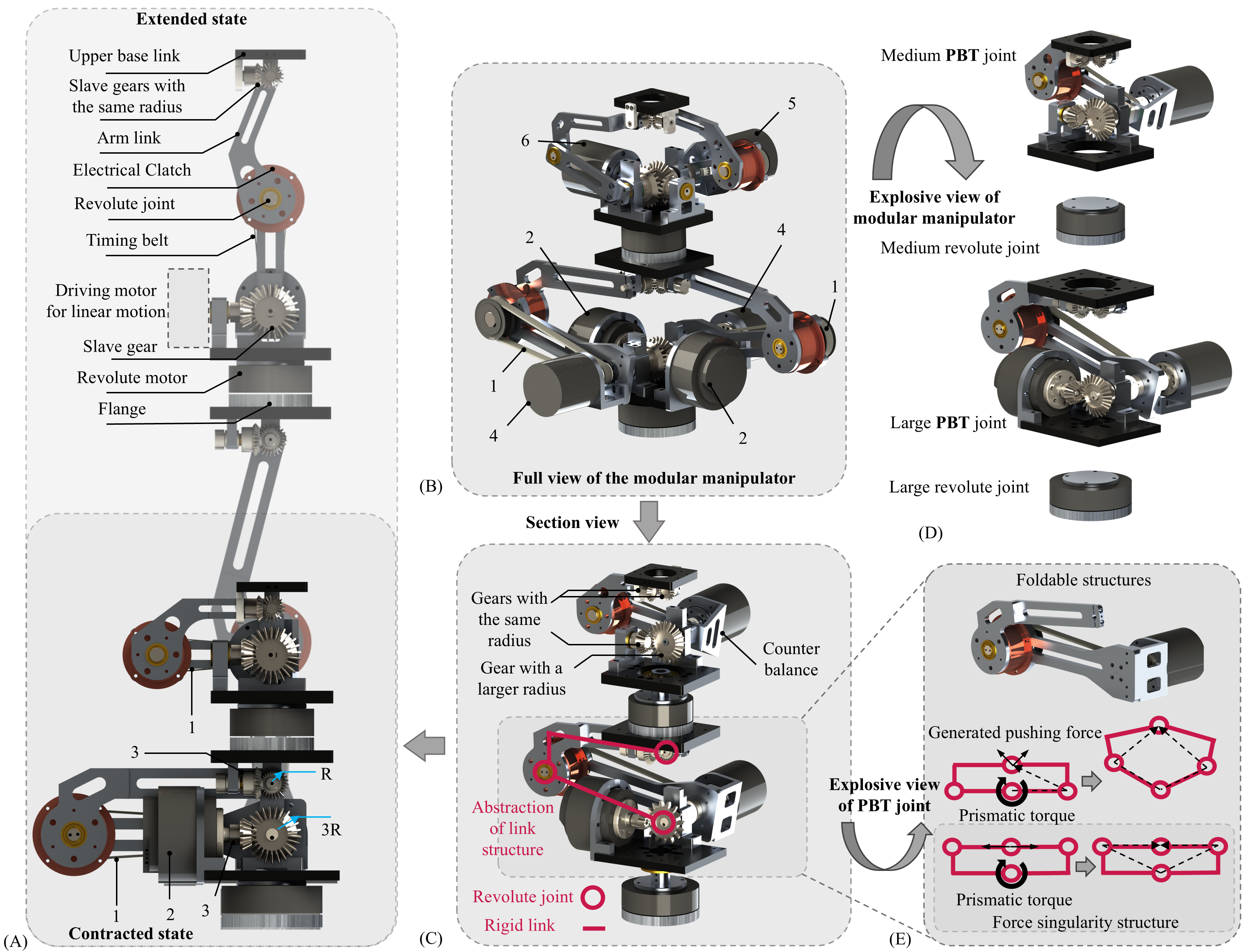}
    \caption{Structure and Assembly of the Two-Module PBT Joint Modular Manipulator
(A) Partial sectional front view of the modular manipulator: 1 represents the timing belt, 2 is the servo motor driving the linear motion of the PBT joint, and 3 represents the driving gear of the PBT joint directly connected to the servo motor. (B) Partial exploded sectional view of the modular manipulator: From bottom to top are the large revolute module, large PBT module, medium revolute module, and medium PBT module. Two potential structures for folding are shown. Force singularity occurs in the structure in the second row. (C) Full view of the modular manipulator: 4 represents the servo motors driving the revolute motion of the PBT joint, 5 represents the servo motor driving the revolute motion of the medium PBT joint, and 6 represents the servo motor driving the linear motion of the medium PBT joint.}
    \label{fig:modular_manipulator}
\end{figure}

\subsection{Direction Maintenance Mechanism}

Although Figs. \ref{fig:Concept}A and \ref{fig:Concept}B intuitively illustrate the relative directions of the coordinates of the PBT joint, translating this concept into practical implementation introduces challenges related to direction maintenance. In traditional joint-link mechanisms, the 
\begin{equation*}
    link \rightarrow joint \rightarrow link
\end{equation*}
architecture allows for seamless assembly in both serial and parallel robotic structures. However, in the PBT joint, multiple links are connected at a single joint, forming a 
\begin{equation*}
\begin{bmatrix}
    link \\
    link \\
    \ldots
\end{bmatrix}
\begin{matrix}
    \searrow \\
    \rightarrow\\
    \nearrow
\end{matrix}
joint 
\begin{matrix}
    \nearrow \\
    \rightarrow\\
    \searrow
\end{matrix}
\begin{bmatrix}
    link \\
    link \\
    \ldots
\end{bmatrix}
\end{equation*}
 architecture. If the motor stator is connected to one link and the rotor to another, it creates a direction maintenance issue, as shown in Fig. \ref{fig:direction_maintain}A. This setup disrupts the endpoint’s direction.

The 3D direction maintenance mechanism is critical for ensuring the stability and precise functionality of the PBT joint in three-dimensional applications. Without it, the connection and extension of PBT joints cannot maintain consistent directionality between modules, leaving the connections floating and rotatable, which severely compromises the robot’s maneuverability. This mechanism effectively addresses the direction maintenance challenge by mechanistically ensuring alignment and withstanding torque along the x, y, and z axes, delivering robust and customizable performance to enhance the joint’s reliability in unstructured and confined environments.

As illustrated in Figure \ref{fig:direction_maintain}B and \ref{fig:direction_maintain}C, the proposed 3D direction maintenance mechanism consists of driving gears, slave gears, base links, and arm links. The driving and slave gears are securely fixed to the base links, and by rotating one or both driving gears (\#1 and \#2), the arm links near the base link open or close, facilitating the linear motion of the PBT joint. The interlocking properties of the differential mechanism ensure high stability and adaptability, making it a reliable solution for dynamic and complex spatial tasks. This design significantly improves the PBT joint’s performance in unstructured environments.

\subsection{Design and Actuation of the PBT Joint}

Figure \ref{fig:modular_manipulator}A and \ref{fig:modular_manipulator}B provide a comprehensive illustration of the PBT joint's structure, including the 3D direction maintenance mechanism with a gear reduction ratio, synchronized servo motor pairs, singularity-free foldable links, and a revolute joint with reduction. These four components are essential for the stable operation of the PBT joint under load. Customized design parameters are summarized in Tab. \ref{table: major parameters}.

The 3D direction maintenance mechanism, in conjunction with synchronized servo motors, enables the linear motion of the PBT joint (Fig. \ref{fig:modular_manipulator}B). Building on the previously introduced 3D direction maintenance mechanism, we adjust the gear reduction ratio by varying the radii of the driving gear and the slave gear. For example, if $R_1:R_2 = 1$, a 90-degree motor rotation results in a $2L$ linear movement of the PBT joint's endpoint. Additionally, we utilize two driving gears to further enhance the load-carrying capability during linear motion. As the direction maintenance mechanism employs a differential gear structure, driving gears can be positioned in any of the other four directions. We select a pair of opposing gears as the driving gears to balance the center of gravity. The servo motors are directly connected to the driving gears through couplings. Notably, the embedded synchronization feature of these servo motors eliminates the need for extensive compliance design or synchronization algorithms, allowing straightforward synchronized control of the two motors.

The singularity-free foldable links are an optimized variation of the closed-loop four-bar linkage structure. This optimization aims to achieve a higher folding ratio without compromising the functionality of the PBT joint (Fig. \ref{fig:modular_manipulator}E). Considering the 3D direction maintenance mechanism occupies the central space and prevents fully overlapping folding, we adopt a stacked link structure. Furthermore, to avoid force singularities at the lower positions, which could hinder linear motion, the foldable links are designed with links \#1 and \#2 as straight links, and links \#3 and \#4 as L-shaped links. This design achieves effective folding while avoiding force singularities.

The revolute joint in the PBT joint comprises an electromagnetic clutch, a synchronous belt reduction mechanism, and a pair of servo motors (Fig. \ref{fig:modular_manipulator}A). During linear motion, the revolute joint acts as a passive joint, with the clutch disengaged to ensure no interference between the motors for linear and revolute motion. When the system transitions to rotational mode (at the farthest linear position), the clutch engages. The synchronous belt transmission provides three key advantages: (1) reducing the inertia of the PBT joint by shifting the main weight closer to the base, (2) introducing a reduction ratio to increase joint torque, and (3) leveraging the compliance of the belt to compensate for any synchronization imperfections in the servo motors.

\begin{table}[t]
\begin{center}
\caption{CUSTOMIZED PARAMETERS OF THE MODULAR MANIPULATOR}
\begin{tabular}{|p{4cm}|p{3cm}|}
    \hline   %第一行线
    \multicolumn{1}{|c}{\centering $Parameters$} &
    \multicolumn{1}{|c|}{\centering $Value$} \\
    \hline  \hline %第一行线
    \multicolumn{2}{|c|}{\centering Large PBT Joint}\\
    \hline
    Reduction ratio of linear motion & 3\\
    \hline
    Reduction ratio of revolute motion & 2\\
    \hline
    Length of link 1 & 16 ($cm$)\\
    \hline
    Length of link 3 & 20 ($cm$)\\
    \hline
    Self-Weight & 5 ($kg$)\\
    \hline
    Linear Payload & 9 ($kg$) \\
    \hline
    \multicolumn{2}{|c|}{\centering Medium PBT Joint}\\
    \hline
    Reduction ratio of linear motion & 3\\
    \hline
    Reduction ratio of revolute motion & 2\\
    \hline
    Length of link 1 & 10 ($cm$)\\
    \hline
    Length of link 3 & 12 ($cm$)\\
    \hline
    Self-Weight & 2.2 ($kg$) \\
    \hline
    Linear Payload & 4 ($kg$) \\
    \hline
    \multicolumn{2}{|c|}{\centering Small PBT Joint (PBT Wrist)}\\
    \hline
    Reduction ratio of linear motion & 2\\
    \hline
    Reduction ratio of revolute motion & 1\\
    \hline
    Length of link 1 & 5 ($cm$)\\
    \hline
    Length of link 3 & 7 ($cm$)\\
    \hline
    Self-Weight & 0.5 ($kg$) \\
    \hline
    Linear Payload & 0.8 ($kg$) \\
    \hline
\end{tabular}
\end{center}
\vspace{-0.2cm}  %????????????
\label{table: major parameters}
\end{table}

\subsection{Assembly of Multi-PBT Joints for Modular Manipulator}

The modular manipulator consists of multiple PBT joints and revolute joints, where revolute joints are independent rotary motors (Fig. \ref{fig:modular_manipulator}B and Fig. \ref{fig:modular_manipulator}C). The assembly of the modular manipulator allows for customization in two aspects: the sequence of joints and the size of the joints. These customizations are tailored based on task requirements and torque demands.

From the base to the end effector, the structure includes a revolute joint, a PBT joint, a revolute joint, and a PBT joint (Fig. \ref{fig:modular_manipulator}D). Besides this configuration, other sequences may be more suitable for different tasks. For instance, a single PBT joint or a combination of a PBT joint and a revolute joint can serve as a dexterous wrist (Fig. \ref{fig:intro}B). A series of PBT joints can enable planar manipulator operations, while a custom configuration of PBT and revolute joints can create an extendable exoskeleton to potentially address current challenges in exoskeleton teleoperation. To accommodate varying torque demands, we provide three sizes of PBT joints: large, medium, and small. Their specific parameters are summarized in Table\ref{table: major parameters}. Taking the modular manipulator in this study as an example, the large and medium modules are used as the base and elbow, respectively, to meet the torque requirements of different positions (Fig. \ref{fig:modular_manipulator}A).

\section{Kinematics and Manipulability Analysis}
\label{sec:kinematics}

This section establishes a comprehensive kinematics analysis for the modular manipulator. We first propose a detailed inverse kinematics (IK) solution for a two-unit modular unit arm, then showcase the extended reachability set of PBT joints with different modes.

\subsection{Inverse Kinematics}
\label{subsec:ik}

Given the target end effector position $p_t$ and $M$ convex obstacles $\{\mathcal{O}_m\}_{m=1}^M$ in the task space, IK aims to find a collision-free joint configuration that reaches $p_t$ \cite{kucuk2006robot}. With two serially arranged modular units, there are four possible motion modes as shown in Fig. \ref{fig:Concept}. Let $l_1$ and $l_2$ respectively be the arm link length of the lower and upper units. Use $t_1$ and $t_2$ to denote the thickness of the base revolute motors of the lower and upper units. $\phi_1$ and $\phi_2$ denotes the joint pose of the two base revolute joints, whereas $\theta_1$ and $\theta_2$ denotes the central revolute joint poses. When a unit is fully stretched, $\theta$'s are zero. For linear motion, the $\theta$ value should be non-negative.

\begin{figure*}
\centering
    \includegraphics[width=\linewidth]{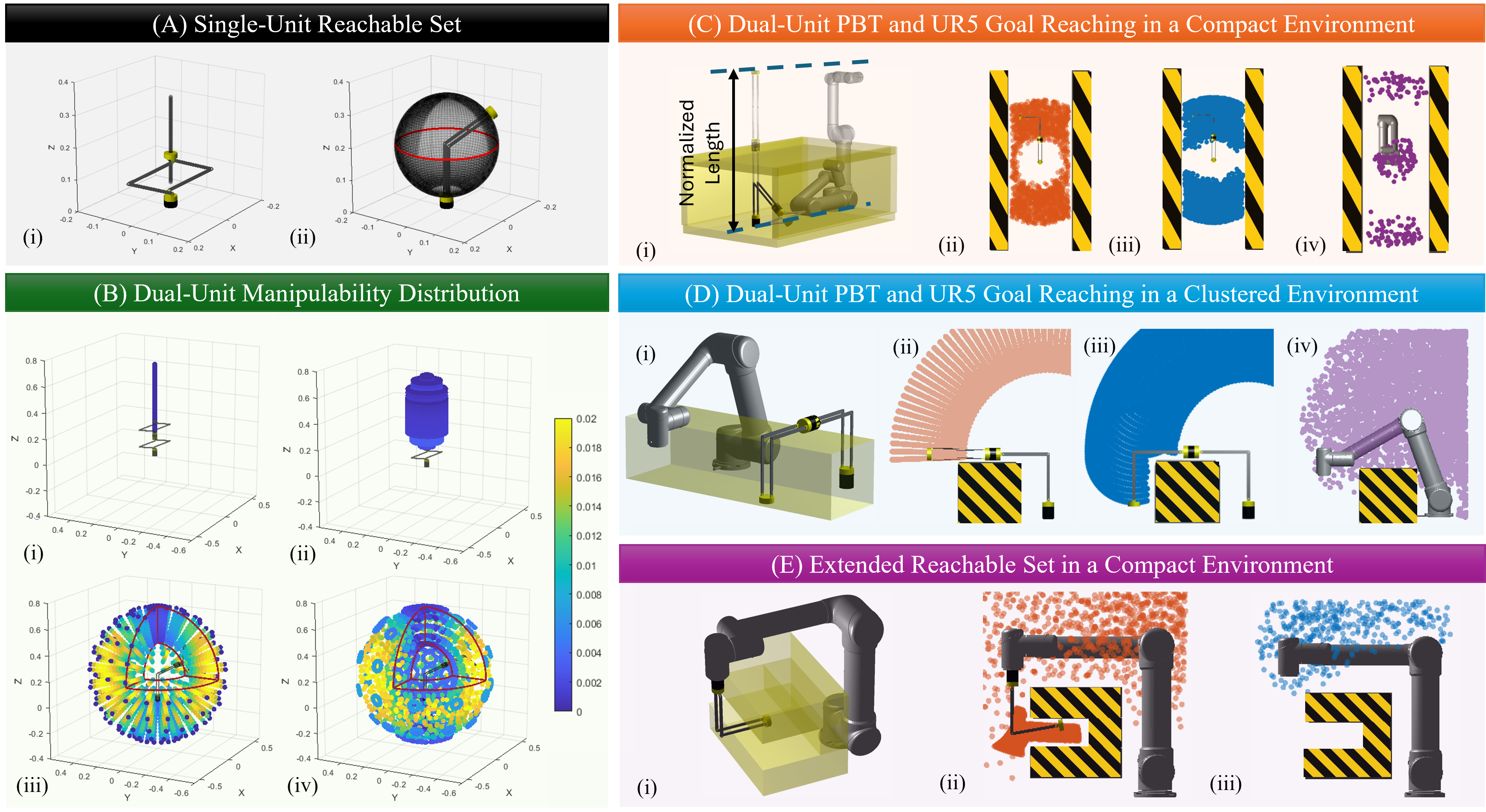}
\caption{\textbf{Reachability and Manipulability Analysis of PBT Joints} (A) Reachability set of a single PBT joint: (i) and (ii) correspond to the prismatic and bending modes, respectively.
(B) Translational Manipulability Distribution in Task Space of dual-unit PBT: (i)-(iv) correspond to the PP, PB, BP, and BB modes defined in Fig. \ref{fig:Concept}.
\textbf{Comparing different motion modes with a conventional UR5 arm} in an compact (C) and a clustered (D) environments.
(i) Environment visualization, the dual-unit PBT and UR5 are scaled to a normalized stretched length. (ii)-(iV) show the BP mode, BB mode and UR5 workspaces. The scatter plots are generated from 30,000 randomly sampled joint configurations, excluding collisions. 
\textbf{Extended conventional robot's workspace with PBT units}
(E) The reachability set in a compact environment is extended with a PBT joint mounted on a UR5. (i) visualizes the environment, while (ii) and (iii) show workspaces with and without the mounted PBT joint. In (ii), sampling density is increased in the cavity area to better capture reachability.}
\label{fig:reachable}
\end{figure*}

%\vspace{-2em}

\subsubsection{PP mode} This is a trivial motion mode where the end-effector can move only up and down linearly in the task space. The IK solution lies in this mode only when $p_t$ is right above the modular manipulator's base.
\subsubsection{BB mode} When both units are operating bending modes, the modular manipulator can be equivalently viewed as four cylindrical links connected serially with revolute joints. There are thorough studies and several commercially ready software packages for IK of this type of structure, such as the MATLAB robotic system toolbox etc. Therefore, we also skip the detailed discussions of this mode here.
\subsubsection{PB mode} In this mode, the Cartesian position of the end-effector can be mapped from the joint poses.
\begin{equation}
\begin{cases}
    x=l_2\sin(\theta_2)\cos(\phi_1+\phi_2) \\
    y=l_2\sin(\theta_2)\sin(\phi_1+\phi_2) \\
    z=t_1+t_2+2l_1\cos(\frac{\theta_1}{2})+l_2\cos(\theta_2)
\end{cases}
\end{equation}

The result is straightforward with arithmetic.
\begin{equation}
\begin{cases}
    \phi_1+\phi_2=\arctan2(y, x) \\
    \theta_1=2\arccos(\frac{z-t_1-t_2-\sqrt{l_2^2-x^2-y^2}}{2l_1}) \\
    \theta_2=\arcsin(\frac{\sqrt{x^2+y^2}}{l_2})
\end{cases}
\end{equation}

Since the first unit is operating linear motion, $\theta_1\geq 0$, so there is only one possible choice of $\theta_1$ and two possible choices of $\theta_2$. Combining with the redundancy of $\phi_1$ and $\phi_2$, this gives the freedom to choose collision-free joint configurations. An intuitive and efficient way is iterating through discrete values of $\phi_1$ and the two values of $\theta_2$ to find a configuration with the furthest distance from obstacles $\{\mathcal{O}_m\}_{m=1}^M$. The redundancies in $\phi_1$, $\phi_2$ and $\theta_2$ are sometimes critical when avoiding collisions.

\subsubsection{BP mode} In this mode, the Cartesian position of the end-effector can be mapped from the joint poses.
\begin{equation}
\begin{cases}
    x=(l_1+t_2+2l_2\cos(\frac{\theta_2}{2}))\sin(\theta_1)\cos(\phi_1) \\
    y=(l_1+t_2+2l_2\cos(\frac{\theta_2}{2}))\sin(\theta_1)\sin(\phi_1) \\
    z=t_1+l_1+(l_1+t_2+2l_2\cos(\frac{\theta_2}{2}))\cos(\theta_1)
\end{cases}
\label{eq:fk}
\end{equation}

The result is also straightforward with arithmetic
\begin{equation}
\begin{cases}
    \phi_1=\arctan2(y, x) \\
    \theta_1=\arcsin(\frac{\sqrt{x^2+y^2}}{\sqrt{x^2+y^2+(z-t_1-l_1)^2}})\\
    \theta_2=2\arccos(\frac{\sqrt{x^2+y^2+(z-t_1-l_1)^2}-l_1-t_2}{2l_2})
\end{cases}
\end{equation}

Similarly, there is also only one possible choice of $\theta_2$ and two possible choices of $\theta_1$. $\phi_2$ is not participated in the forward kinematics (\ref{eq:fk}), which, analogous to the LR mode, leaves freedom to select collision-free joint configurations.

Although a solution for two-unit systems is provided here, efficient IK methods for modular manipulators with more units remains a challenge due to the curse of dimensionality. Further research is needed to address this issue.

\subsection{Reachability and Manipulability Analysis}
This part analyzes the reachability and manipulability of PBT joints, both as an attached tool of a collaborative robot arm and a two-unit modular manipulator \cite{bicchi1995mobility}.

Fig. \ref{fig:reachable}A shows the reachable set of a single PBT joint, from which we can see that the P mode has a reachable set of a line and the B mode has sphere reachable set. Fig. \ref{fig:reachable}B visualizes the translational manipulability of a dual-unit PBT distributed in the end effector's reachable positions. Translational manipulability describes the required effort to drive the robot towards all possible task-space directions at a given configuration. It is quantified here following the definition in \cite{yoshikawa1985manipulability}
\begin{equation}
    \sigma(J)=\sqrt{\det(J(q)J^T(q))}=\sqrt{s_1s_2\dots s_n}
\label{eq:manipulability}
\end{equation}
where $J(q)$ is the translational Jacobian at configuration $q$, and $s_1\dots s_n$ are eigenvalues of matrix $J(q)J^T(q)$. Under PP mode, the modular manipulator can only move linearly up and down, so its reachable set is a line with zero manipulability. BP and BB modes has similar reachable set and manipulability distribution in obstacle-free environment, and has a larger volume in comparison to the workspace of PP and PB modes. In general, mode transformation makes only a minor improvement when moving in a free space.

The advantages of mode transformation becomes evident in the presence of obstacles. Prismatic mode allows PBT joints to shrink its dimensions, enhancing flexibility in compact environments, while bending mode enables the end-effector to navigate around obstacles, offering greater dexterity in clustered scenarios. Since no general quantitative measure for collision-free dexterity exists, a qualitative analysis is provided. Figure \ref{fig:reachable}C shows performance in a narrow tunnel. BB mode creates a split workspace, limiting transitions, while BP mode maintains connectivity for better flexibility. The UR5, despite a similar size, twists excessively and has the most fragmented workspace. Figure \ref{fig:reachable}D illustrates performance in clustered environments. BP mode cannot reach beyond obstacles, while BB mode achieves a larger collision-free workspace, demonstrating superior dexterity. The UR5 has a smaller reach on the other side comparing to the BB mode.

Fig. \ref{fig:reachable}E demonstrates the capability of PBT joints to extend a robot's workspace in a compact environment. The PBT joint allows manipulation in a cavity area whose opening is opposite to the robot's base. The robot end-effector is unable to enter the cavity area without the assistance of a PBT joint.

\subsection{Inverse Dynamics}

The inverse dynamics refers to the problem of determining the joint torques $\tau$ given the robot motion $\theta, \dot{\theta}, \ddot{\theta} $. The recursive Newton-Euler algorithm is efficient to solve the serial robot manipulators. As the bending mode is similer to serial manipulator, we focus on the dynamics of prismatic mode of PBT joint.

As the PBT joint has a symmetrical structure, we use Lagrangian method
\begin{equation}
    \frac{d}{d t}\left(\frac{\partial L}{\partial \dot{\theta}}\right)-\frac{\partial L}{\partial \theta}=\tau
\end{equation}
to obtain its dynamics.

Assuming $^{\rm w}v_{i-1}, ^{\rm w}\omega_{i-1}$ represent the linear and angular velocity from the previous link in the world frame, the velocity of the center of mass and angular velocity of the PBT joint is
\begin{equation}
\begin{cases}
    ^{\rm w}v_i = ^{\rm w}v_{i-1} + ^{i-1}R_{i} \begin{bmatrix}
 0\\
 0\\
 -lsin\frac{\theta}{2}\dot{\theta}
\end{bmatrix} \\
    ^{\rm w}\omega_i = ^{\rm w}\omega_{i-1}
\end{cases}
\end{equation}
where $^{i-1}R_{i}$ represents the rotation matrix from frame $i-1$ to the PBT frame $i$. The kinetic and potential energy of the PBT joint is
\begin{equation}
\begin{cases}
    K_i=\frac{1}{2} m_i v_i^T v_i+\frac{1}{2} \omega_i^T I_i \omega_i \\
    U_i=m_i g^T r_{i}
\end{cases}
\end{equation}
where $I_i=\text{diag}(\frac{1}{2}m_il^2,\frac{1}{2}m_il^2\cos^2\frac{\theta}{2},\frac{1}{2}m_il^2\sin^2\frac{\theta}{2})$ is the inertial matrix. It is straightforward to obtain the Lagrangian operator on the PBT joint as
\begin{align}
    & \tau = \nonumber\\
    & m_i l^2 \sin^2\frac{\theta}{2} \ddot{\theta} + 
      m_i l^2 \sin\frac{\theta}{2} \cos\frac{\theta}{2} \dot{\theta}^2 
      - \frac{1}{2} m_i l^2 \sin^2\frac{\theta}{2} \dot{\theta}^2 \nonumber \\
    & + m_i (^{\rm pbt} \dot{v}_{i-1})^T
      \begin{bmatrix}
          0 \\ 0 \\ -l \sin\frac{\theta}{2}
      \end{bmatrix}
      - \frac{1}{2} m_i l g^T \cdot ^{i-1}R_i \begin{bmatrix}
          0 \\ 0 \\ \sin\frac{\theta}{2} \end{bmatrix} \\
    & - \frac{1}{2} \omega_{i-1}^T \frac{\partial I_i}{\partial \theta} \omega_i
    + m_i (^{\rm pbt} v_{i-1})^T
      \begin{bmatrix}
          0 \\ 0 \\ \frac{\dot{\theta} l}{2} (\sin\frac{\theta}{2} - \cos\frac{\theta}{2})
      \end{bmatrix} \nonumber.
\end{align}
where the $^{\rm pbt}\dot{v}_{i-1}$ represents the velocity of previous link in the PBT joint's frame. It's obvious that when the base is static and $\theta=\pi$, the maximum required torque to lift the PBT joint is $\frac{1}{2}m_igl$. As analyzed in Fig. \ref{fig:modular_manipulator}B, if the applied torque is insufficient, lifting the joint may be impossible in certain kinematic configurations.

\section{Experimental Validation}
\label{sec:experiments}

In the experimental section, we mainly demonstrate two use cases based on the PBT joints. One is the PBT wrist, and the other is the modular manipulator. The configurations of their experimental setups and the explanations and discussions of the demonstrations are included in this section.

\begin{figure*}
\centering
    \includegraphics[width=\linewidth]{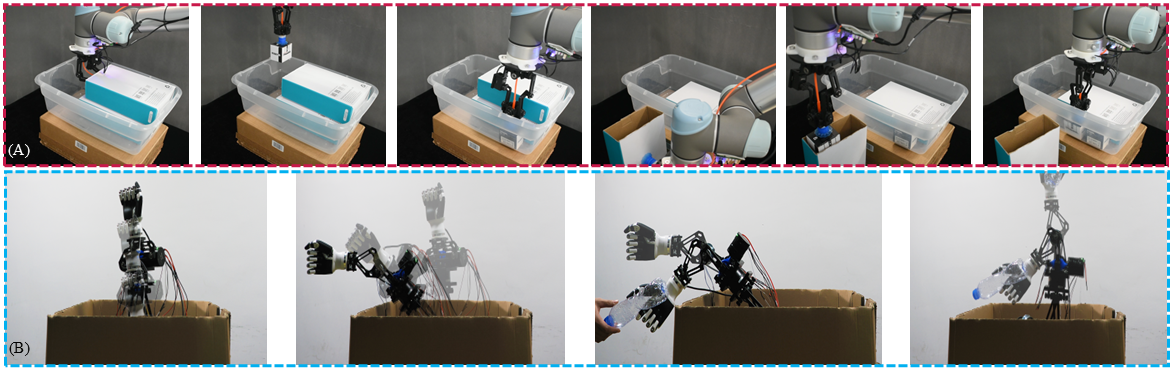}
\caption{Demonstration of Single-Joint PBT Wrist and Two-Joint Modular Manipulator
(A) PBT Wrist Demonstration: A pick-and-place process is performed using the PBT wrist integrated on a UR5 robotic arm. The sequence involves the PBT wrist reaching into a large box to pick out a smaller box and stacking it in a narrow space. Subsequently, the large box is picked, reoriented, and further utilized for additional box-picking tasks. Finally, the picked box is stacked in the narrow space.
(B) Modular Arm Demonstration: The two-joint PBT modular manipulator showcases its foldable and reconfigurable capabilities. The arm unfolds from within a box, reaches a specified pose, grasps a plastic bottle, and then folds back into the box.}
\label{fig:exp}
\end{figure*}

\subsection{Demonstration of The One Module PBT Wrist}

The PBT wrist is a small-sized PBT joint (Fig. \ref{fig:direction_maintain}). It also uses the same link configuration (Fig. \ref{fig:modular_manipulator}E) to avoid singularities in force during lifting and lowering motions, with its parameters as shown in Tab. \ref{table: major parameters}. Unlike the previously shown PBT joint, the wrist's revolute joint is not driven by a synchronous belt but rather by directly mounting a Dynamixel servo motor (M226) on the revolute axis, which directly drives the revolute motion. Thus, the reduction ratio of the revolute joint is 1. Another identical Dynamixel servo motor is installed on the orientation-maintaining mechanism, driving the wrist's linear motion through a gear structure with a reduction ratio of 2. Both motors are connected to the PC via a servo driver board, enabling simple control (Fig. \ref{fig:intro}B and \ref{fig:intro}C) \cite{zhou2024design}.

To showcase the dexterity of the PBT wrist, we attached a suction cup at the wrist's end to achieve pick-and-place operations in confined spaces (Fig. \ref{fig:exp}A). When vertical space is limited, the foldable and bendable structure at the end of the wrist allows effective grasping, manipulation, and placement while also increasing the variety of grasping directions.

\subsection{Demonstration of Two Module Manipulator}

The structure of the modular manipulator is shown in Fig. \ref{fig:modular_manipulator}. It uses two types of motors: the x6-40 and x4-24 motors (MyActuator Limited, Suzhou). A pair of x6-40 motors are used in the large PBT joint to drive prismatic motion, while the revolute motion of the large PBT joint is driven by a pair of x4-24 motors. Additionally, the two joints of the medium PBT joint are each driven by a single motor, with the prismatic joint and revolute joint both driven by x4-24 motors. Lastly, the revolute motors at the base of each joint are x6-40 motors. These motors are internally equipped with motor drivers, allowing direct communication with the PC for simple control of the motor groups, including synchronized control (simultaneous operation) of the motor pairs.

The motion of the module manipulator is presented in Fig. \ref{fig:exp}B. The large PBT joint provides three degrees of freedom: bottom rotation and prismatic-bending dual-modal motion. Electromagnetic clutches manage the revolute joints, disengaging during prismatic motion for passive following and engaging for revolute motion when aligned. Demonstrations showcase telescopic and grasping motions within a box, emphasizing the manipulator’s high telescoping ratio and dexterity in confined space.

\section{Conclusion and Future Work}
In this paper, we introduced the Prismatic-Bending Transformable (PBT) Joint, a novel robotic joint that integrates multi-directional bending, rotation, and elongation/contraction within a compact, modular, and reconfigurable architecture. Inspired by scissor mechanisms and designed with directional maintenance consideration, the PBT Joint addresses limitations of conventional joint-link chain manipulators—particularly the lack of integrated dexterity and structural transformability. We demonstrated its versatility through two representative applications: a single-joint PBT wrist extension for existing robotic arms and a dual-joint foldable modular manipulator. Both configurations exhibited enhanced dexterity, expanded workspace coverage, and improved obstacle avoidance in confined and cluttered environments. Comprehensive validation through detailed design, mechanical optimization, kinematic and dynamic analysis, and experimental testing confirmed the PBT Joint’s effectiveness in enabling adaptable manipulation across a broad range of task scenarios.

Future work will focus on developing advanced control and motion planning strategies tailored to the PBT joint’s unique transformable capabilities, enabling precise, adaptive manipulation in cluttered and dynamic environments. Additional directions include scaling to multi-joint configurations, miniaturization for fine manipulation, and integration with humanoid and autonomous systems. Enhancing payload capacity and energy efficiency will further support deployment in demanding industrial applications.



\bibliographystyle{IEEEtran}
\bibliography{main}
%\begin{thebibliography}{1}


%\end{thebibliography}

\begin{IEEEbiography}
[{\includegraphics[width=1in,height=1.25in]{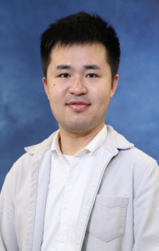}}]
{Jianshu Zhou} (Member, IEEE/ASME) received his Ph.D. in Mechanical Engineering from The University of Hong Kong in 2020. He is currently a postdoctoral scholar at the Mechanical Systems Control Lab, Department of Mechanical Engineering, University of California, Berkeley. Prior to this, he served as a research assistant professor at The Chinese University of Hong Kong. His research interests include robotics, mechatronics, grasping and manipulation, novel actuators, and dexterous hands.
\end{IEEEbiography}

\begin{IEEEbiography}
[{\includegraphics[width=1in,height=1.25 in]{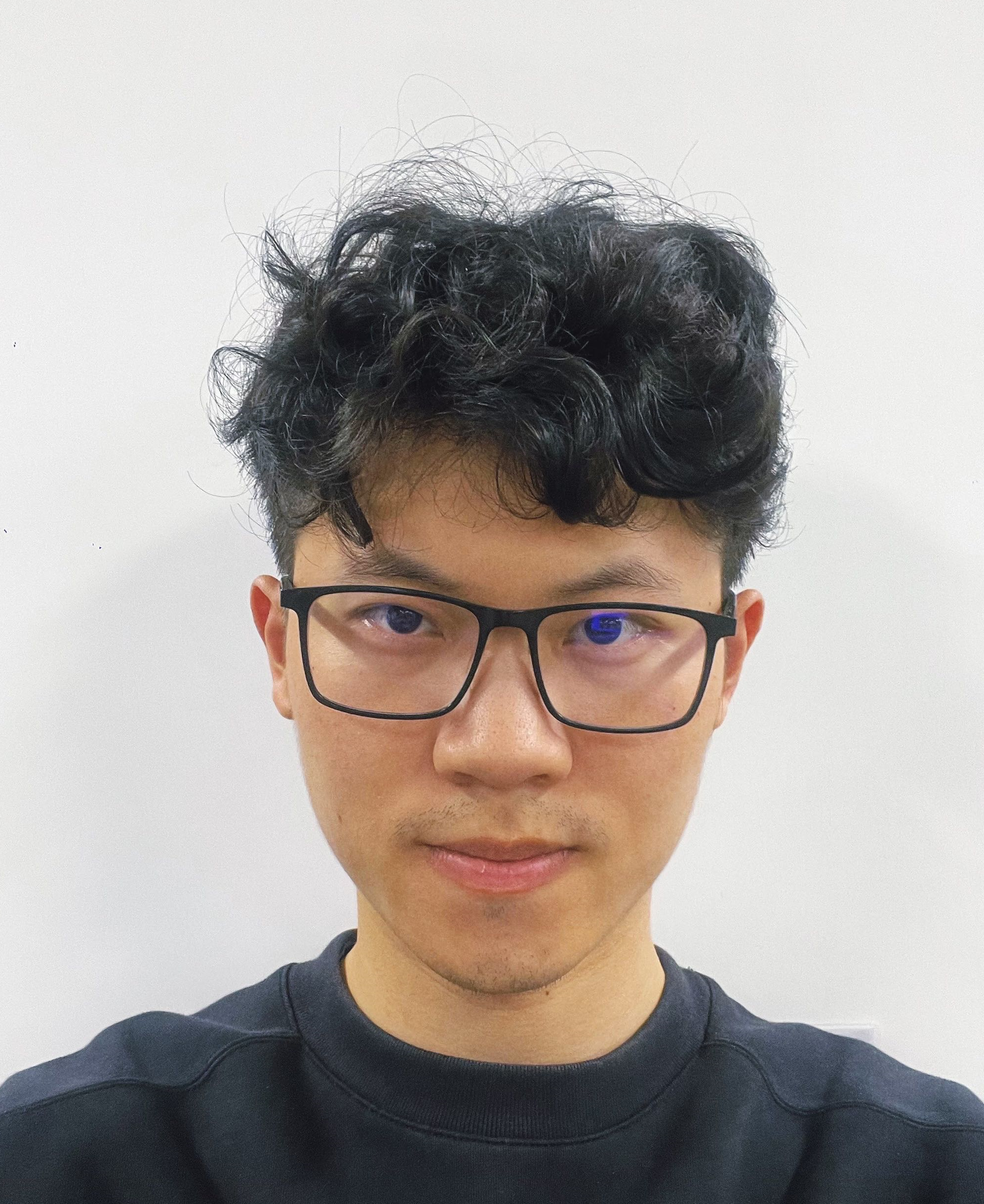}}]
{Junda Huang} received the B.Eng. degrees from the University of Science and Technology of China (USTC) in 2020. Currently, he is pursing the Ph.D. degree at the CUHK. His research interest includes dexterous hand, robotic grasping and manipulation, teleoperation, and imitation learning.
\end{IEEEbiography}

\begin{IEEEbiography}
[{\includegraphics[width=1in,height=1.25in]{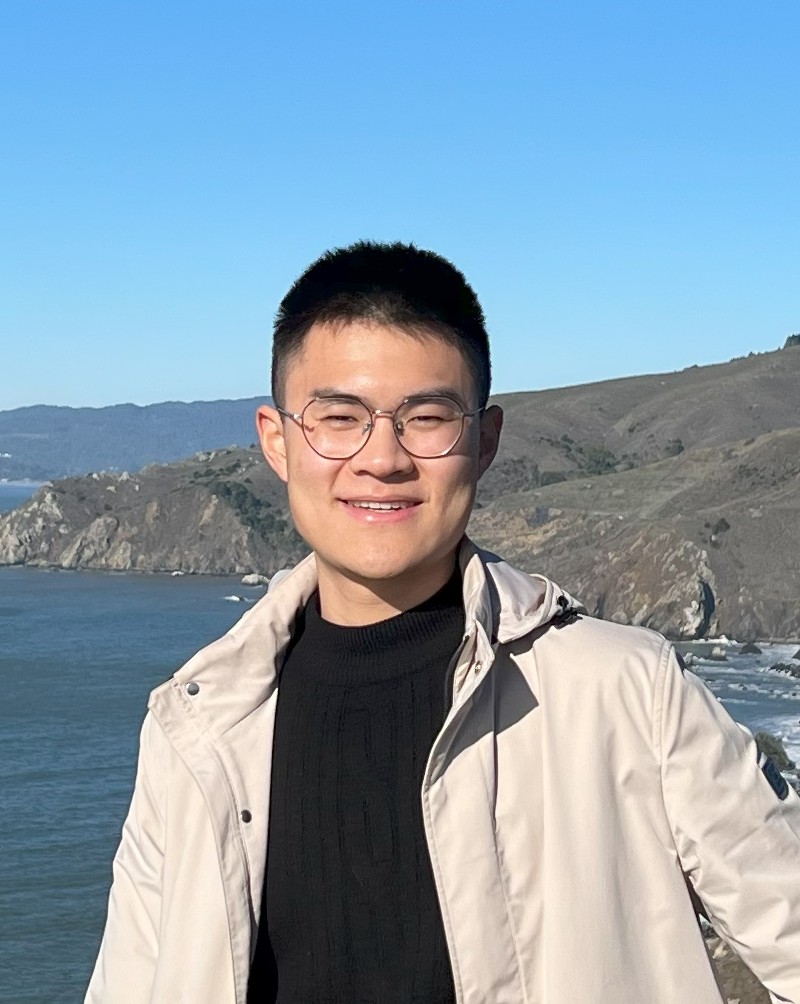}}]
{Boyuan Liang} received the B.Sc. (Hons) degree in Applied Mathematics from National University of Singapore in 2021. He is currently pursuing Ph.D. degree in Mechanical Engineering at University of California, Berkeley. His research interest lies in robotics, mechatronics systems, teleoperation and contact modeling.
\end{IEEEbiography}

\begin{IEEEbiography}
[{\includegraphics[width=1in,height=1.25in]{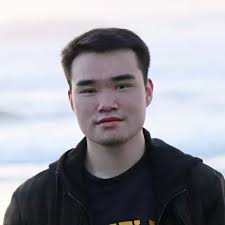}}]
{Xiang Zhang} received his B.Eng. degree in 2019 from the University of Science and Technology of China (USTC) and his Ph.D. degree in 2024 from the University of California, Berkeley. He is now with FANUC Advanced Research Laboratory, FANUC America Corporation, USA. His research interests include robot modeling, control, and manipulation.
\end{IEEEbiography}

\begin{IEEEbiography}
[{\includegraphics[width=1in,height=1.25in]{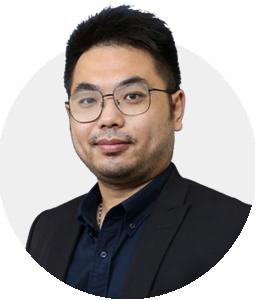}}]
{Xin Ma} (Member, IEEE) received the B.Eng. and Ph.D. degrees in mechanical and electronic engineering from Dalian University of Technology, Dalian, China, in 2011 and 2017, respectively.,From 2017 to 2019, he was with The Chinese University of Hong Kong, Hong Kong, as Research Post-Doctoral Fellow. From 2019 to 2021, he was with Purdue University, West Lafayette, IN, USA, as Post-Doctoral Fellow. He is currently a Research Assistant Professor with The Chinese University of Hong Kong.
\end{IEEEbiography}

\begin{IEEEbiography}
[{\includegraphics[width=1in,height=1.25in]{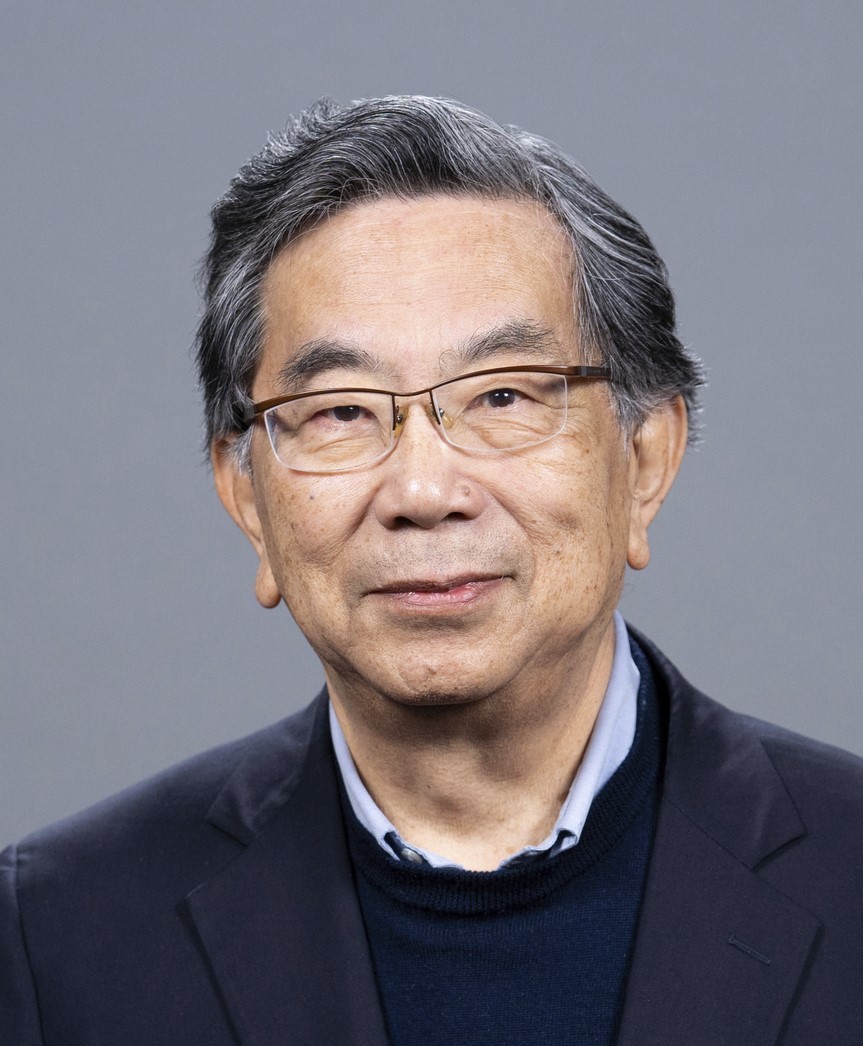}}]
{Masayoshi Tomizuka} (Life Fellow, IEEE/ASME) was born in Tokyo, Japan, in 1946. He received the B.S. and M.S. degrees in mechanical engineering from Keio University, Tokyo, Japan, in 1968 and 1970, respectively, and the Ph.D. degree in mechanical engineering from the Massachusetts Institute of Technology, Cambridge, MA, USA, in February 1974.,In 1974, he joined the Faculty of the Department of Mechanical Engineering, University of California at Berkeley, Berkeley, CA, USA, where he is currently the Cheryl and John Neerhout, Jr., Distinguished Professorship Chair. At UC Berkeley, he teaches courses in dynamic systems and controls. From 2002 to 2004, he was the Program Director of the Dynamic Systems and Control Program of the Civil and Mechanical Systems Division of NSF. His current research interests are optimal and adaptive control, digital control, signal processing, motion control, and control problems related to robotics, machining, manufacturing, information storage devices, and vehicles.,Dr. Tomizuka was the Technical Editor of the ASME Journal of Dynamic Systems, Measurement and Control, J-DSMC (1988–1993), Editor-in-Chief of the IEEE/ASME Transactions on Mechatronics (1997–1999), and Associate Editor for the Journal of the International Federation of Automatic Control, and Automatica. He was the General Chairman of the 1995 American Control Conference, and was the President of the American Automatic Control Council (1998–1999). He is a Life Fellow of the ASME and a Fellow of the International Federation of Automatic Control (IFAC) and the Society of Manufacturing Engineers. He was the recipient of the Best J-DSMC Best Paper Award (1995, 2010), DSCD Outstanding Investigator Award (1996), Charles Russ Richards Memorial Award (ASME, 1997), Rufus Oldenburger Medal (ASME, 2002), John R. Ragazzini Award (AACC, 2006), Richard E. Bellman Control Heritage Award (AACC, 2018), Honda Medal (ASME, 2019), and Nathaniel B. Nichols Medal (IFAC, 2020). He is a member of the National Academy of Engineering.

\end{IEEEbiography}


%\end{IEEEbiography}


%\newpage



%\vfill


\end{document}